\documentclass[conference]{IEEEtran}
\usepackage{times}

\usepackage[numbers]{natbib}
\usepackage{multicol}
\usepackage[bookmarks=true]{hyperref}

\usepackage{graphics} 
\usepackage{amsmath} 
\usepackage[noend]{algpseudocode}
\usepackage{subfig}
\usepackage{algorithmicx,algorithm}
\usepackage{multirow}
\usepackage{hyperref}
\usepackage{float}
\usepackage{booktabs}
\usepackage{mathrsfs}
\usepackage{bbm}
\usepackage{tabularx}
\usepackage{color}
\usepackage{graphicx}
\usepackage{caption}

\usepackage{threeparttable}
\usepackage{graphicx}
\usepackage{amssymb}
\usepackage{xspace}

\usepackage[bookmarks=true]{hyperref}

\begin{document}

\title{\textbf{\texttt{SInViG}}: \\A Self-Evolving Interactive Visual Agent for Human-Robot Interaction
}


\author{\authorblockN{Jie Xu$^{\alpha, \beta}$, Hanbo Zhang$^{\beta, *}$, Xinghang Li$^{\gamma, \beta}$, Huaping Liu$^{\gamma}$, Xuguang Lan$^{\alpha, *}$, Tao Kong$^{\beta}$}
\authorblockA{$^\alpha$ Xi'an Jiaotong University, $^\beta$ ByteDance Research, $^\gamma$ Tsinghua University\\
* Correspondence to: zhb@bytedance.com, xglan@mail.xjtu.edu.cn}
}


%

\newcommand{\figref}[1]{Fig.\ref{#1}}
\newcommand{\tabref}[1]{Table.\ref{#1}}
\newcommand{\secref}[1]{Section.\ref{#1}}
\newcommand{\apref}[1]{Appendix}
\newcommand{\sinvig}{SInViG\xspace}
\newcommand{\rtxt}[1]{\textcolor{red}{#1}}
\newcommand{\btxt}[1]{\textcolor{blue}{#1}}
\newcommand{\subfig}[1]{\textit{#1}}
\newcommand{\boxbin}[1]{$<$\textit{BIN}\_$#1$$>$}

\maketitle

\begin{abstract}
Linguistic ambiguity is ubiquitous in our daily lives\footnotetext{\textbf{$\dag$ Work done during an internship at ByteDance Research}}.
Previous works adopted interaction between robots and humans for language disambiguation.
Nevertheless, when interactive robots are deployed in daily environments, there are significant challenges for natural human-robot interaction, stemming from complex and unpredictable visual inputs, open-ended interaction, and diverse user demands.
In this paper, we present \sinvig, which is a self-evolving interactive visual agent for human-robot interaction based on natural languages, aiming to resolve language ambiguity, if any, through multi-turn visual-language dialogues.
It continuously and automatically learns from unlabeled images and large language models, without human intervention, to be more robust against visual and linguistic complexity.
Benefiting from self-evolving, it sets new state-of-the-art on several interactive visual grounding benchmarks.
Moreover, our human-robot interaction experiments show that the evolved models consistently acquire more and more preferences from human users.
Besides, we also deployed our model on a Franka robot for interactive manipulation tasks.
Results demonstrate that our model can follow diverse user instructions and interact naturally with humans in natural language, despite the complexity and disturbance of the environment.

\end{abstract}

\IEEEpeerreviewmaketitle

\section{Introduction}

Robots are gradually entering our daily lives as intelligent helpers.
Seamlessly integrating into human environments requires them to act upon natural language instructions.
Nevertheless, ambiguity is ubiquitous in language expressions.
Therefore, handling ambiguity is significant for robots in human-robot interaction (HRI).
One natural way of addressing visual-language ambiguity is to interact for clarification, which presents multiple significant challenges:
\begin{itemize}
    \item Robustness to complex and unpredictable visual inputs.
    \item Robustness to open-ended interaction.
    \item Robustness to diversified users.
\end{itemize}
Concretely, visual inputs may contain open-vocabulary objects or visual concepts, occlusions, or cluttered backgrounds.
Unrestricted interaction with humans requires robots to possess advanced cognitive abilities, such as contextual reasoning, disambiguation, robustness to unlearned concepts, and flexibility to adapt to a wide range of communication styles and preferences.
More importantly, individual users may exhibit distinct preferences for modes of interaction, necessitating a multi-modal approach in human-robot communication systems.
For instance, individuals with hearing impairments may rely exclusively on gesture-based interactions. 
Furthermore, users may naturally interact in multi-modality, such as speech accompanied by gestures.
Therefore, the interaction must accommodate this diversity to ensure usability for all users.
Extensive visual datasets \cite{changpinyo2021conceptual, ordonez2011im2text, zhu2023minigpt4, das2017visual, de2017guesswhat, openimages, invig, yu2016modeling, kirillov2023segment, schuhmann2022laion, gadre2023datacomp, kakaobrain2022coyo-700m, shao2019objects365} are crucial for robust visual representation learning \cite{bengio2013representation, uelwer2023survey}, while large language models \cite{du2021glm, touvron2023llama, touvron2023llama2, team2023gemini, chowdhery2023palm} can be excellent teachers for language-based interaction.
A natural question arises: \textit{Can we harness the extensive visual data and the impressive power of large language models to resolve the problem of interactive visual grounding considering the mentioned challenges?}


In this paper, we propose \sinvig, namely Self-evolving INteractive VIsual Grounding, an automatic and self-evolving learning system, for robustness when facing the complexity of the open world.
Our system takes two external knowledge sources: millions of unlabeled image data for visual robustness and a large language model for language robustness.
It iteratively and automatically learns from the existing datasets while labeling new data using the improved model with the help of large language models (LLMs).
We have shown that, by training on the self-labeled and LLM-polished interaction data, the model can continuously improve its performance and robustness against challenging examples.
Moreover, it learns from the LLMs and asks more diversified, accurate, and informative questions when facing ambiguity.
The iteration converges after three rounds of learning, resulting in a highly robust system against visual complexity, open-ended interaction, and diversified users, for open-world HRI.

We have conducted experiments on diversified benchmarks, including the standard visual-language benchmarks,  HRI and scoring with humans, and real-robot interactive manipulation tasks, to comprehensively verify the robustness of our proposed method.

For validation on visual-language tasks, experimental results demonstrated that \sinvig achieved new state-of-the-art performance with a clear margin on the InViG \cite{invig} benchmark.
Moreover, results also demonstrate that the model after self-evolving outperforms the models from previous iterations by a clear margin, verifying the efficacy of the self-evolving iteration.

For experiments of human-robot interaction, we follow TiO \cite{xu2024towards} to validate the performance on its HRI bench, which includes a selected set of 150 challenging scenarios for interactive visual grounding.
In this part, we recruit volunteers to interact with \sinvig.
Results show that \sinvig achieves an overall success rate of 74\%, exceeding the baseline by a margin of 6\%.
Moreover, human scoring shows that \sinvig after self-evolving is much more preferred compared to baselines.

For real-robot interactive manipulation tasks, we have deployed \sinvig on Kinova Gen 3 robot.
In this part, we have validated its performance first on interactive grasping tasks \cite{shridhar2018interactive, shridhar2020ingress, yang2022interactive}.
Results show that our method can achieve an overall grasp success rate of 82.2\% with highly complex visual observations and diversified instructions.

One main contribution of this work is that we propose a close-loop and efficient self-evolving system for interactive visual grounding, which is also promising to transfer to other interactive robotic systems.
By taking advantage of this self-evolving loop, it is highly robust to nearly arbitrary open-world interactive scenarios both visually and linguistically.
This paper also presents a million-scale, LLM-enhanced interactive visual grounding dataset that can effectively train a policy to ground target objects visually and interactively. 
We hope that it can facilitate the progress of research on vision-based HRI. 


\section{Related Work}

\subsection{Interactive Visual Grounding}
\textit{Interactive Visual Grounding} (IVG) \cite{shridhar2020ingress} is stemmed from Visual Grounding (e.g. \cite{nagaraja2016modeling, yu2016modeling, yu2018mattnet, qiao2020referring, deng2021transvg}).
Given the possible ambiguity in the initial natural language description, it is to ground the target object visually by interacting with the users.
It has been actively explored recently, especially for multi-modal human-robot interaction \cite{goodrich2008human}.
In robotics, recent works usually rely on an integrated system to resolve this problem, most probably including a visual grounding model to connect the visual perception and natural language instructions and an interactive policy to further generate language responses \cite{whitney2017reducing, shridhar2018interactive, mees2021composing, yang2022interactive}.
Interactive grounded dialogues are also demonstrated to be useful to online improvements of perceptual modules \cite{thomason2019improving}.
To handle IVG in clutter, Zhang et al. \cite{zhang2021invigorate} integrate POMDP \cite{kaelbling1998planning} with visual relationships \cite{zhang2018visual} for interactive grasping in object-blocking scenarios and show that planning with uncertainty can improve robustness due to visual occlusion.
Recently, thanks to the advances in open-vocabulary visual-language models like CLIP \cite{radford2021learning}, it is feasible to build interactive systems to solve IVG with open-world objects \cite{mo2023towards}.
Related to IVG, Visual question generation (VQG) has also been explored as pure visual-language problems \cite{whitney2017reducing, de2017guesswhat, mora2016towards, mostafazadeh2017image, mostafazadeh2016generating, kottur2019clevr, matsumori2021unified}.
For example, the dataset named GuessWhat?! \cite{de2017guesswhat} is proposed to solve the problem of iteratively asking judgment questions and guessing the target object given an image as the input.
Following this dataset, a series of end-to-end interactive models are proposed and show promising results  \cite{tu2021learning, pang2020visual}.
Nevertheless, it restricts the possible answers from the users to `Y', `N', or `N/A', which makes the interaction less flexible and sometimes confusing.

\subsection{LLMs for Vision-based Interaction}

Large Language Models (LLMs) have demonstrated remarkable proficiency in language understanding and inference, prompting recent efforts to extend their capabilities to multi-modal domains, encompassing vision, text, audio, and other modalities~\cite{li2023vision, alayrac2022flamingo, liu2023visual, zhang2021vinvl, zhou2022conditional, zhu2023minigpt, ranasinghe2023perceptual}. To enhance the vision-based interaction capabilities of LLMs, various vision tasks, such as image captioning~\cite{vinyals2015show, lu2017exploring, li2021robotic}, vision question answering~\cite{antol2015vqa, das2017visual}, and visual grounding~\cite{fukui2016multimodal, yu2016modeling, yu2018mattnet}, are incorporated into the training phase. For instance, \citet{peng2023kosmos} introduce patch embedding for top-left and bottom-right patches, serving as representations for bounding boxes, thereby enabling visual grounding for LLMs. Subsequent advancements extend LLMs to encompass pixel-level visual grounding~\cite{rasheed2023glamm, wang2023cogvlm, chen2023minigpt}, incorporating segmentation and referred object masks. To leverage the comprehensive comprehension and inference abilities of LLMs in handling ambiguous tasks, a reasoning segmentation task is proposed, facilitating the inference of mentioned objects and their precise pixel-level locations~\cite{lai2023lisa}. In the field of robotics, recent studies leverage visual grounding capabilities for text-conditioned affordance, generating heatmaps of object components that align with task requirements~\cite{qian2024affordancellm}.

While the aforementioned works successfully establish connections between natural language and visual concepts, they often overlook the integration of human feedback, e.g., human-robot interaction. 
In contrast, our proposed method takes a significant step forward by enhancing the interaction capacity of LLMs through a self-evolving algorithm. This approach enables LLMs to disambiguate and accurately locate target objects during conversations with humans.

\begin{figure*}[t!]
    \centering
    \includegraphics[width=\textwidth]{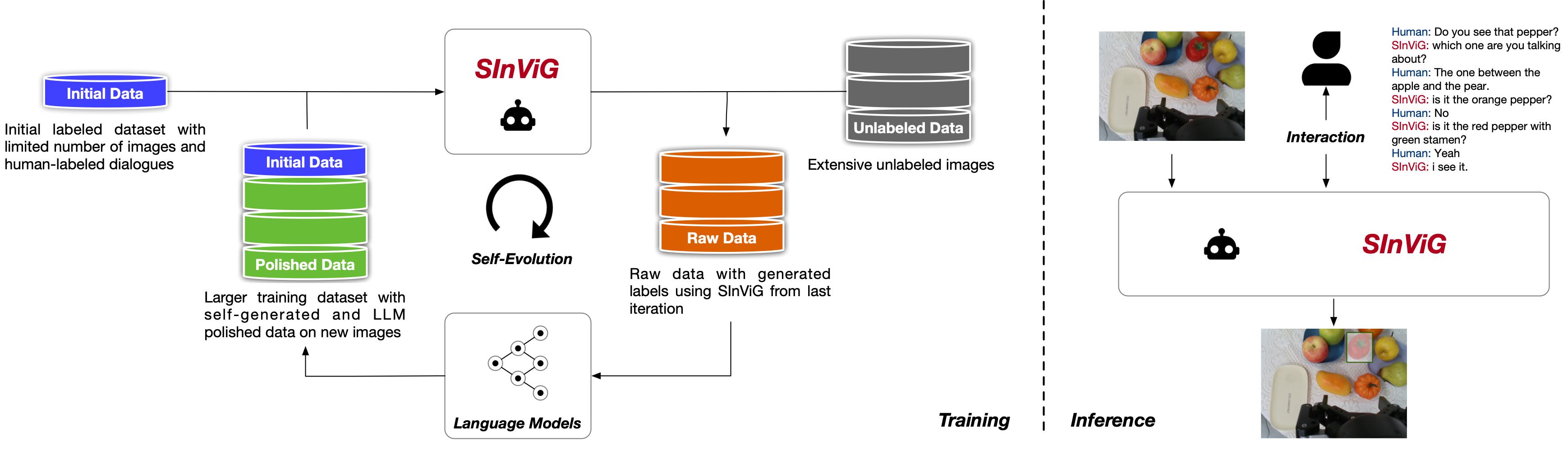}
    \caption{Overview of \sinvig. \sinvig is a self-evolving model for interactive visual grounding. Left: self-evolving training iteration; Right: the demonstration of inference steps for \sinvig.}
    \label{fig:intro}
\end{figure*}

\subsection{Iterative Self-Improved Learning}
To improve the model's performance without additional data labeling, multiple training paradigms are proposed, including reinforcement learning and self-supervised learning. Reinforcement learning \cite{li2017deep, henderson2018deep, hester2018deep, wang2020reinforcement, haarnoja2018soft} requires a subtly designed reward function to evaluate the rollout of the policy, which is especially hard in language-conditioned tasks due to the diversity of natural language. 
Similarly, early works in self-supervised learning design novel pretext tasks like jigsaw puzzling~\cite{noroozi2016unsupervised}, image colorization~\cite{zhang2016colorful} and rotation classification~\cite{gidaris2018unsupervised}.  With the introduction of contrastive learning, self-supervised learning brings excellent performance gain to multiple downstream tasks such as text retrieval, object detection, and image classification. The core idea of contrastive learning is to increase the similarity between the semantically similar sample pairs while decreasing the similarity between the opposite pairs. 

Rather than relying on a handcrafted reward to enlarge the training database or construct an initial representation space for the training samples, concurrent works \cite{huang2022large, shridhar2023art} introduce the LLMs as an external data source and propose the idea of self-improving, which iteratively enlarges the training data by the feedback from LLMs and improves the performance of the model. \citet{yuan2024self} regards LLM as a judger that predicts the reward of the generated responses and adds the preference pairs to the training data. Inspired by constrastive learning, \citet{wang2023improving} utilizes LLM to generate positive and hard-negative samples and adopts InfoNCE loss to learn the text embeddings for documents. Instead of judging the quality of synthetic data by the LLM, an additional binary classifier is trained to select generated data and enlarge the dataset for the next-step fine-tuning~\cite{singh2023beyond}. In the field of robotics, RoboCat~\cite{bousmalis2023robocat} constructs a training loop to first finetune the model with human-annotated data and further self-generate more data to enlarge the training set to boost the performance of the model. 

\section{Problem Formulation}
\label{sec:formulation}

Interactive visual grounding (IVG) is the task of grounding the target objects given ambiguous descriptions through multi-turn interaction. Typically, it takes an image $I$ and an initial expression $E$ as inputs.
In each interaction turn $t$, the robot needs to decide whether to ask a disambiguation question $q_t$ or provide the final decision, in general, an object $o_t=(x_{min}, y_{min}, x_{max}, y_{max})$ as the output, indicating the coordinates of the upper-left corner and the bottom-right corner of the corresponding bounding box.

To train the model, each sample $(E, I, \hat{H}, \hat{o})$ consists of an initial expression $E$, an image $I$, a disambiguation dialog $\hat{H}_T=\{(\hat{q}_i, \hat{a}_i\}_{0}^{T}$ and a labeled object $\hat{o}=(\hat{x}_{min}, \hat{y}_{min}, \hat{x}_{max}, \hat{y}_{max})$, where $\hat{\cdot}$ means the ground truth, and $\hat{a}_t$ represents the answer of question $\hat{q}_t$.
At turn $t$, we directly concatenate the dialogue history before $t$ as contextual information to condition the outputs: $\hat{H}_t=(\hat{q}_0, \hat{a}_0, ..., \hat{q}_t, \hat{a}_t)$.
Let $\theta$ denote the parameters of our model, in each dialogue turn, it directly predicts the question to be asked $q_t$ and the most probable target object $o_t$ given the previous dialogue and visual input: $(q_t, o_t)=f_{\theta}(I_t, \hat{H}_t)$.
We follow previous works \cite{de2017guesswhat, invig, xu2024towards} to call the predictor of $q_t$ the Questioner and the predictor of $o_t$ the Guesser.
Besides, to facilitate automatic evaluation on interactive benchmarks, we need an Oracle to play the role of human users: $a_{t+1}=f_{\phi}(q_{t+1}, I_t, \hat{H}_t, \hat{o})$, where $\phi$ means the parameters of the Oracle, $a_t$ means the predicted answer of $q_t$.
In our implementation, we share the trainable parameters of $\theta$ and $\phi$ following previous work \cite{xu2024towards}.
Note that the Oracle will not be used during human-robot interaction or real-robot deployment.

During inference, in each turn $t$, except for $q_t$ and $o_t$, the model also needs to decide whether the information is enough to locate the target.
To address this issue, we integrate an auxiliary task into our model to predict whether the dialog should stop $s_t \in (Yes, No)$, with a fixed prompt \textit{Is it clear?} as shown in \tabref{table:prompt}.
In each turn, the model will be first invoked with this auxiliary task to predict $s_t=f_{\theta}(I_t, \hat{H}_t)$.
If the $s_t$ is $Yes$, it will predict the target object.
Otherwise, it will ask a question and wait for the answer from the user.


\section{\sinvig}
\label{sec:sinvig}

\subsection{Overview}
\label{sec:overview}

\sinvig is a self-evolving interactive model that continuously and automatically improves robustness against the complexity of visual observations and open-ended language expressions, with extensive unlabeled image data from language models, respectively.
As shown in \figref{fig:intro}, it starts with a human-labeled dataset and a supervised model.
In each iteration, it generates new labels based on unlabeled images and the model from the last iteration automatically.
The generated labels are then polished using the large language models \cite{zhao2023survey}, hopefully to absorb wider and richer text distributions to improve the ability of interaction.
The polished data is then merged with the training data and used to train \sinvig for the next iteration.

Our self-evolving approach is based on the following assumptions:
\begin{itemize}
\item An initial interactive model with reasonable performance. In this paper, the initial model is from supervised learning with a human-labeled dataset.
\item An accessible large language model (e.g. ChatGPT \cite{ouyang2022training}, or LLaMa 2 \cite{touvron2023llama, touvron2023llama2}).
\item A large collection of unlabeled images. If they do not include bounding boxes for objects, an object detector is mandatory to generate object proposals.
\end{itemize}


\subsection{\sinvig Network}
\label{sec:sinvig_net}

We follow TiO \cite{xu2024towards} to design our network architecture.
Concretely, it includes a unified tokenizer for texts and integers (e.g. the coordination of bounding boxes), a ResNet \cite{he2016deep} vision encoder, and a transformer backbone.
All three tasks including the Questioner, Guesser, and Oracle are formulated in a unified framework.
We will introduce the details in this section.

\paragraph{Unified Tokenizer}
Our network is designed to be a unified network with shared parameters for the Questioner, Guesser, and Oracle.
Different roles will be distinguished using different instructions, as shown in \tabref{table:prompt}.
The Questioner and Oracle are both text generation tasks, yet the Guesser is to predict object locations.
To unify the output space, we follow OFA \cite{wang2022ofa, xu2024towards} and build a unified tokenizer for joint training of all three tasks.
Concretely, our tokenizer is built upon the BERT \cite{chen2019bert} tokenizer, including an additional set of 1000 integer tokens: \boxbin{i}, where $i$ is an integer and $i\in [0, 1000)$.
Each bounding box $(x_{min}, y_{min}, x_{max}, y_{max})$ is normalized with the image width and height and then mapped to one of these integers.
Hence, all three tasks can be modeled with token sequences for auto-regressive prediction.

\paragraph{Vision Encoder}
The input image of \sinvig is first scaled to 512$\times$512 and then fed into the vision encoder to get image tokens.
We adopt ResNet-152 \cite{he2016deep} as the backbone of our vision encoder. 
The output features are shaped as 32$\times$32$\times$2048.
It is processed by a linear projection layer and results in 1024 image tokens.
They are directly concatenated with text embeddings for auto-regressive sequential modeling.

\paragraph{Auto-Regressive Transformer} 
We adopt the encoder-decoder backbone from OFA \cite{wang2022ofa, xu2024towards} for the auto-regressive tasks.
Our encoder network takes the concatenated sequences, including both image tokens and text tokens, as input.
It outputs joint embeddings, which are then fused to the decoder layers using cross-attention.
The decoder is responsible for generating outputs auto-regressively.
We adopt a model consisting of 24 encoder layers and 12 decoder layers in our experiments.

\begin{table}
\caption{Prompt Examples}
\label{table:prompt}
\begin{center}
\begin{tabularx}{1\columnwidth}{l@{\hspace*{5pt}}l@{\hspace*{5pt}}l@{}}
\toprule
Role & Task & Prompt  \\
\midrule
Questioner & Asking & Be helpful, and ask for clarification if unsure.\\ 
Guesser & Locating & Be helpful, and output bounding box only. \\
Oracle & Answering & Be helpful, and answer questions.\\
- & Captioning & What do you see?\\
- & Stopping & Is it clear?\\
\bottomrule
\end{tabularx}
\end{center}
\end{table}


\paragraph{Training Loss}
Benefiting from the unified formulation of multiple tasks, our training loss of the Guesser and Questioner follows a simple cross-entropy loss of next-word prediction:
\begin{equation}
    L=-\frac{1}{N}\sum_{N}log(\hat{w}_{l}|\hat{w}_{<l}; I_t, \hat{H}_t)
\end{equation}
where $\hat{w}$ represents a token from the target token sequence and $N$ is the length of the target sequence.
Each target sequence of the Guesser $\hat{o}=(w_1, w_2, w_3, w_4)$ consists of 4 integer tokens, and each target sequence of the Questioner $\hat{q_t}=(w_1, ..., w_N)$ is the tokenization of a target question.
Similarly, for the Oracle, the loss is:
\begin{equation}
    L=-\frac{1}{N}\sum_{N}log(\hat{w}_{l}|\hat{w}_{<l}; q_t, I_t, \hat{H}_t, \hat{o})
\end{equation}
with the conditions of the question $q_t$ and the target object $o$.
Note that all conditions will be tokenized and concatenated as the input of the encoder model.















\subsection{Data Generation}
\label{sec:data_gen}

In this section, we introduce the data generation of each self-evolving iteration in detail.
In summary, we directly utilize \sinvig to generate the labels for unlabeled images for the model training of the next self-evolving iteration.
Therefore, to launch the self-evolution, we need an extensive set of unlabeled images (\secref{sec:data_gen}(a)) and an initial \sinvig model trained using human-labeled data to generate new dialogue data (\secref{sec:data_gen}(b)).


\paragraph{Data Source}
Since our dialogues are generated based on specific and meaningful regions in the images, we need to generate object proposals using object detectors (e.g. \cite{zhou2022detecting}) if the labels of object locations are not available.
As an alternative, we can also directly use object detection or segmentation datasets (e.g. \cite{kirillov2023segment, openimages, shao2019objects365}).
In this paper, we follow the latter choice.
Concretely, our data generation is based upon 3.1 million images during our self-evolving iteration, with 2 million coming from the SAM dataset \cite{kirillov2023segment} and 1.1 million from the OpenImages dataset \cite{openimages}.

\paragraph{Dialogue Generation}
To generate dialogues, we directly use the interaction among the Questioner, Guesser, and Oracle of \sinvig from the last self-evolving iteration.
For each image, at the beginning of dialogue generation, we randomly select an object as the target.
The conversation starts with the Oracle, which is to give an initial description of the object.
Then the conversation continues between the Oracle and the Guesser in a turn-talking manner.
In each turn, after the Oracle gives an answer, we invoke the stop predictor introduced in \secref{sec:formulation} and \tabref{table:prompt} to decide whether to stop for visual grounding.
If yes, the conversation stops, and the Guesser outputs an object bounding box, which is used to compare to the original selected one.
If the intersection over union (IoU) is larger than 0.5, this data point will be saved.


\subsection{Data Polishing with LLMs}

To enhance the interactive ability of \sinvig, we utilize large language models for dialogue polishing.
It stemmed from our observation that the generated dialogues are semantically less diverse compared to the training dataset.
If we continue to iterate directly using the generated data, it leads to a degradation of language comprehension and generation capabilities.

In this phase, firstly, we use \sinvig from the last iteration to generate a detailed caption for each data point.
The caption is directly generated by prompting our model to follow the prompt \textit{What do you see?} as shown in \tabref{table:prompt}.
The detailed caption provides the LLM with more detailed visual information, which helps to generate semantically richer dialogue content.
Then, we prompt the LLM based on the caption and the raw dialogue generated using \sinvig to:
\begin{itemize}
\item 1) List key points of the dialogue;
\item 2) List several potential scenarios for the dialogue;
\item 3) Based on a randomly selected scenario, modify the dialogue by asking questions to better fit the scenario;
\item 4) Simplify the dialogue.
\end{itemize}
The dialogues generated from 3 and 4 are the enriched version and simplified version of the original dialogue, respectively.
During training, we randomly select one of them as the label of interaction data.



\begin{table}
\caption{Performance on InViG Benchmark}
\label{table:invig}
\begin{center}
\begin{tabularx}{1\columnwidth}{l@{\hspace*{35pt}}c@{\hspace*{35pt}}c@{}}
\toprule
 & SR of MT-VG & SR of IVG \\
\specialrule{0em}{1pt}{1pt}
\midrule
ViLBert \cite{tu2021learning} & 55.1\% & - \\
\specialrule{0em}{1pt}{1pt}
X-VLM \cite{zeng2022xvlm} & 59.7\% & 40.1\% \\
\specialrule{0em}{1pt}{1pt}
TiO \cite{xu2024towards} & 77.1\% & 78.1\% \\
\midrule
\specialrule{0em}{1pt}{1pt}
\sinvig-R2 & \textbf{77.4\%} & \textbf{80.9\%} \\
\bottomrule
\end{tabularx}
\end{center}
\end{table}

\section{Experimental Settings}

\subsection{Implementation Details}

\sinvig model is initialized with the pretrained weights from TiO \cite{xu2024towards}, with approximately 900 million trainable parameters.
During the self-evolving training iteration, \sinvig is trained with 8 A100 GPUs, each equipped with 80GB of memory.
Benefiting from BF16 \cite{kalamkar2019study}, our model can be trained with a total batch size of 112.
Detailed hyperparameters of training can be found in \apref{ap:hyperparam}.

\subsection{Datasets}

Except for the automatically generated dataset, we found that the incorporation of other visual-language datasets can essentially improve the final performance, including visual grounding \cite{yu2016modeling}, image caption \cite{ordonez2011im2text, zhu2023minigpt4}, object detection \cite{openimages}, and visual dialogues \cite{das2017visual, de2017guesswhat, liu2023visual}.
Hence, we have trained two series of \sinvig models, one only with data of interactive visual grounding, namely \sinvig-HO, and another with a mixed dataset including open-source visual-language datasets and our data, namely \sinvig, where HO means homogeneous and HE means heterogeneous.
The weights of different datasets follow the principle that in each iteration, the automatically generated data will be trained for 1 epoch and the others will be trained for 0.1 epoch.
This empirical setting has been verified in our experiments that the model can converge without overfitting to any dataset.

\subsection{Tasks and Metrics}

We have validated the performance of \sinvig at three levels: 1) standard visual-language benchmarks; 2) realistic human-robot interaction; 3) interactive robotic manipulation tasks.

For the standard visual-language benchmarks, the robot is required to iteratively ask questions to guess the target object $\hat{o}$, provided the initial ambiguous expression $E$ and observation $I$.
To enable automatic evaluation, following previous works \cite{de2017guesswhat, invig}, we take the Oracle model to replace human users for question answering.
Therefore, we can follow InViG \cite{invig} to evaluate 4 different tasks: \textit{multi-turn visual question generation (MT-VQG)}, \textit{multi-turn visual question answering (MT-VQA)}, \textit{multi-turn visual grounding (MT-VG)}, \textit{interactive visual grounding (IVG)}.
For open-ended language-based interaction (MT-VQG, MT-VQA), we follow two series of metrics: 
1) text similarity to the ground truth, including BLEU1 and BLEU4 (\textbf{B1, B4}) \cite{papineni2002bleu}, ROUGE (\textbf{R}) \cite{lin2004rouge}, METEOR (\textbf{M}) \cite{banerjee2005meteor}, CIDEr (\textbf{C}) \cite{vedantam2015cider}, and SPICE (\textbf{S}) \cite{anderson2016spice}, which are widely used in natural language processing \cite{sai2022survey}; 
2) multi-choice top-1 or top-5 recall (\textbf{R@1} and \textbf{R@5}) and mean ranking (\textbf{Rank}) \cite{das2017visual, invig}, which measure the ability of the model to select the correct choice among a set of disturbances.
Intuitively, text similarity gives hints of the ability to generate human-like utterances, while multi-choice recall measures the correctness of responses.
For MT-VG and IVG, we directly validate the success rate (\textbf{SR}), representing the fraction that the model successfully grounds the target given interaction dialogues.
Noticeably, in IVG, the model is tested in a self-play mode: it takes the interaction dialogues generated by the Guessor and the Oracle to ground the target instead of the ground truth dialogues in MT-VG.

\begin{table}[t!]
\caption{Cross Validation on InViG}
\label{table:cross_val}
\begin{center}
\begin{tabularx}{1\columnwidth}{c@{\hspace*{20pt}}|c@{\hspace*{20pt}}|c@{\hspace*{20pt}}|c@{}}
\toprule
  \specialrule{0em}{1pt}{1pt}
Questioner & Guesser & Oracle & SR of IVG (\%)\\
\midrule
\multicolumn{3}{c|}{\sinvig-R0} & 73.6 \\
  \specialrule{0em}{1pt}{1pt}
\hline
  \specialrule{0em}{1pt}{1pt}
\multicolumn{2}{c|}{\sinvig-R0} & \sinvig-R1 & 75.5 \\
  \specialrule{0em}{1pt}{1pt}
\hline
  \specialrule{0em}{1pt}{1pt}
\multicolumn{2}{c|}{\sinvig-R0} & \sinvig-R2 & 78.5 \\
  \specialrule{0em}{1pt}{1pt}
\hline
  \specialrule{0em}{1pt}{1pt}
\multicolumn{2}{c|}{\sinvig-R1} & \sinvig-R2 & 78.4 \\
  \specialrule{0em}{1pt}{1pt}
\hline
  \specialrule{0em}{1pt}{1pt}
\multicolumn{3}{c|}{\sinvig-R2} & \textbf{80.9} \\

\bottomrule
\end{tabularx}
\renewcommand{\arraystretch}{1.5}
\end{center}
\end{table}

\begin{figure*}
    \begin{tabular}{cc}
       \includegraphics[width=0.32\textwidth]{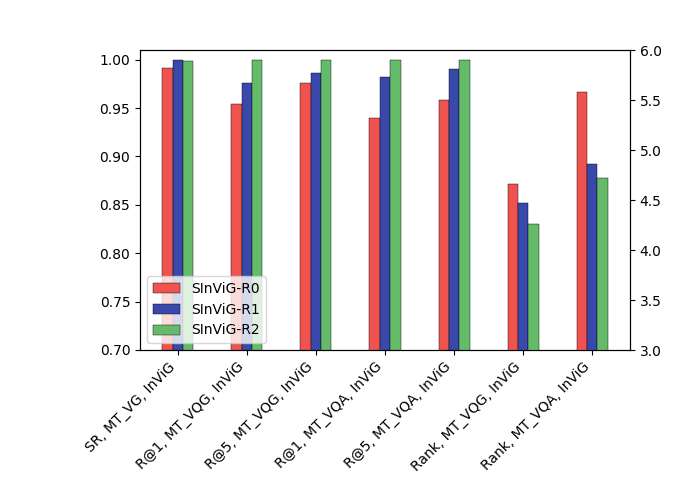}
       \includegraphics[width=0.32\textwidth]{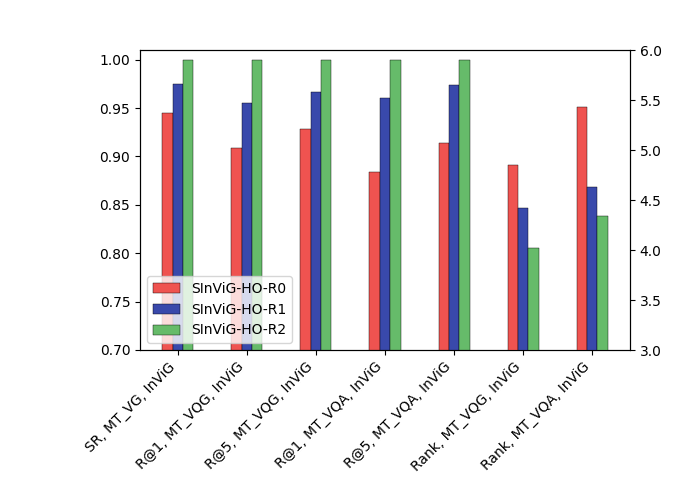}
        &
       \includegraphics[width=0.32\textwidth]{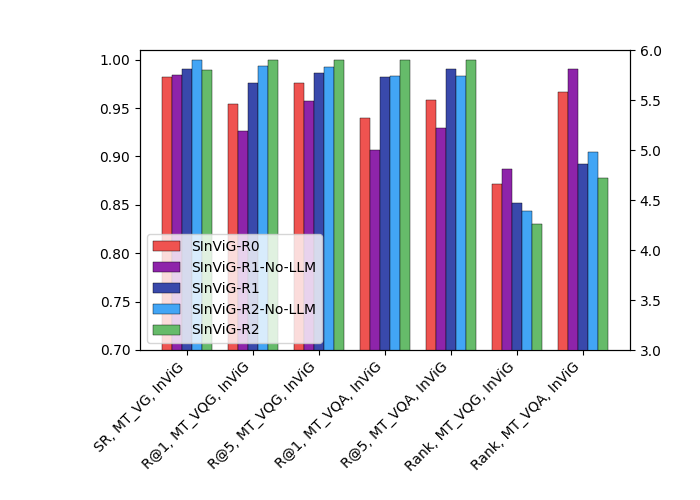}
       \\
       (\subfig a) Self-evolving performance comparison. & (\subfig b) Performance of ablation study.
    \end{tabular}
    \caption{Performance of \sinvig with different self-evolving iteration rounds. (\subfig a) The comparison across models with different rounds of self-evolving iteration. (\subfig b) Ablation study of \sinvig with/without LLM polishment \& additional image sources. Note that lower is better for \textbf{Rank}, and higher is better for other metrics. Metrics except for \textbf{Rank} are normalized (divided) by the maximum value in each group for visualization.}
    \label{fig:invig_exp}
    \vspace{-15pt}
\end{figure*}

For realistic human-robot interaction and interactive robotic manipulation tasks, we directly conduct the experiments with a volunteer to be the question answerer.
The task of the robot is to ask disambiguation questions when facing ambiguity in the initial description and guess the target in the user's mind.
To measure performance, one straightforward metric is the success rate (\textbf{SR}), which is applied in this paper.
Nevertheless, we find that humans can handle minor glitches in the interaction quite effectively.
Hence, only the success rate is not sufficient to reflect the model's capability.
We thereby ask the volunteers to score the models after the interaction to rank the experience of interacting with different models.


\section{Experimental Analysis}

In this section, we aim to answer the following questions:

\begin{itemize}
    \item Where is the upper bound of and how robust is \sinvig? 
    \item What does \sinvig learn from the self-generated data? 
    \item Does \sinvig improve user experience by self-evolving? 
    \item How is \sinvig deployed on real-robot platforms? 
\end{itemize}

\subsection{Where is the upper bound of and how robust is SInViG?}

To answer this question, we iteratively train the model with the self-generated data to improve the performance and generate new data with the improved models following the methods introduced in \secref{sec:sinvig}.

We conduct the validation on the InViG \cite{invig} benchmark, which is designed for the performance evaluation of interactive visual grounding.
Results are demonstrated in \tabref{table:invig}.
We can conclude that:
1) \sinvig performs better than \sinvig-HO, verifying the efficacy of additional public visual-language data during the model training.
2) \sinvig achieves better performance on InViG compared to previous state-of-the-art performance.
We also elaborate on the iterative improvements of models with different rounds of self-evolving, which are shown in \figref{fig:invig_exp}(a). 
The results consistently verify the positive effects of self-evolving iteration, both on \sinvig and \sinvig-HO.
Besides, we provide a detailed analysis of text similarity for MT-VQG and MT-VQA in \apref{ap:txt_sim}, which measures the similarity between the human-labeled annotations and generated texts.
Results reveal that training on LLM-polished data may harm the distribution of texts and achieve lower similarity to human annotation.
Nevertheless, we found that this distribution shift does not necessarily harm the user experience as shown and discussed in \secref{sec:hri_exp}.
We leave a deeper investigation of this observation in our future work.

We also provide a comprehensive comparison with the state-of-the-art multi-modal LLMs including GPT-4 Vision \cite{achiam2023gpt} and Gemini-Pro \cite{team2023gemini} in \apref{ap:mllms_comp}.
We conclude the main observations as follows:
1) For visual grounding, GPT-4 Vision and Gemini-Pro can hardly output object locations. Hence they can only work with the Guesser of \sinvig;
2) For multi-modal interaction, GPT-4 vision and Gemini-Pro are more powerful than \sinvig;
3) For latency, \sinvig can respond in 0.41s on average, yet GPT-4 Vision and Gemini-Pro are much slower (9.22s and 13.72s on average respectively), hence not suitable for real-robot deployment.





\subsection{What does \sinvig learn from the self-generated data?}
\label{sec:ablation}

\begin{figure*}[t!]
    \centering
    \includegraphics[width=0.9\textwidth]{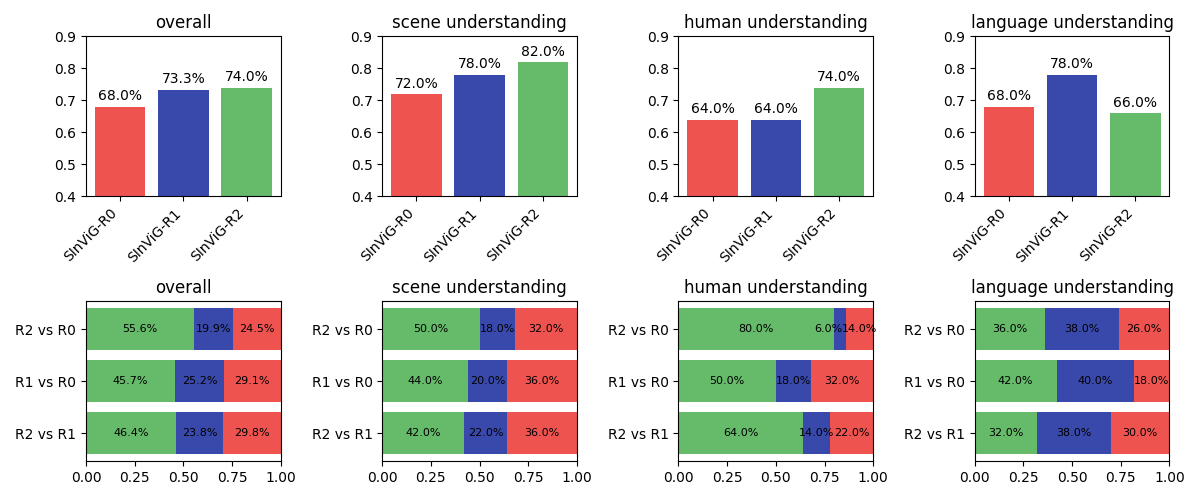}
    \caption{Results of human-robot experiments. Top row: success rate on TiO-HRI bench. Bottom row: scores of human evaluation. Scores are selected from ``better'' (green), ``tie'' (blue), and ``worse'' (red).}
    \label{fig:hri_exp}
\end{figure*}

\begin{figure*}[t]
    \centering
    \includegraphics[width=\textwidth]{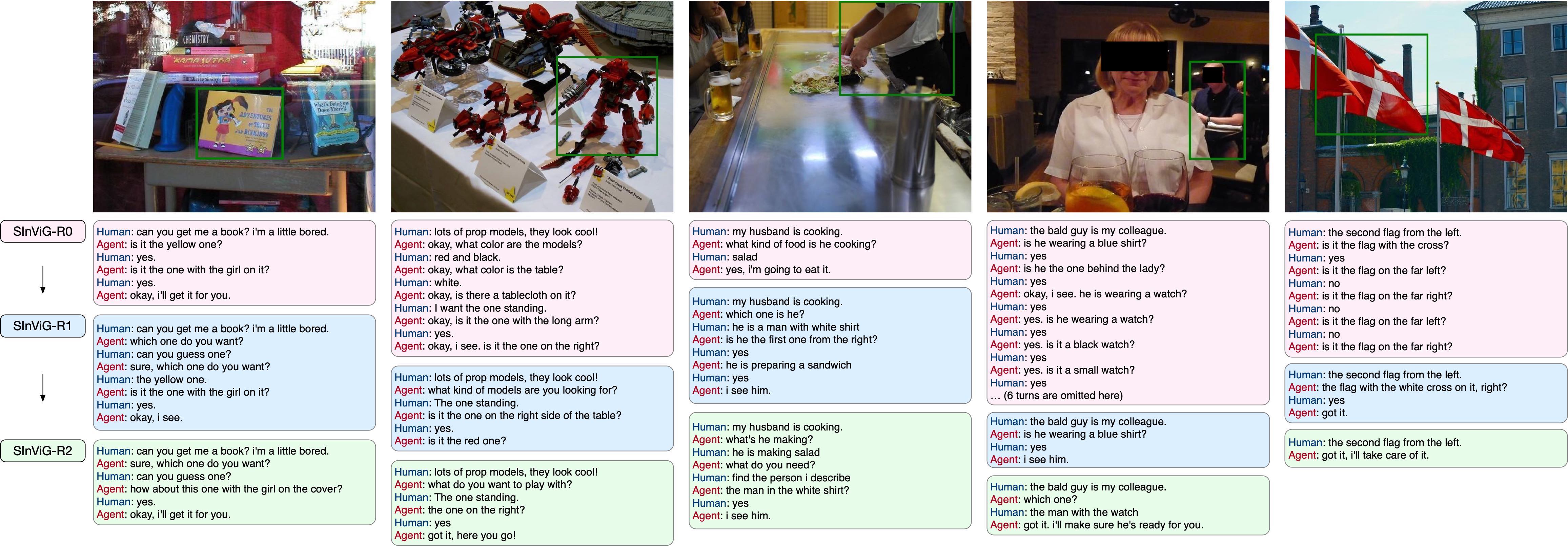}
    \caption{
    Examples of human-robot interaction. 
    We have shown examples from \sinvig-R0 to \sinvig-R2. 
    We can see that the \sinvig-R2 is more efficient and accurate during interaction. 
    It asks fewer redundant questions and follows instructions from the users more properly, benefiting from the LLM-augmented dialogues.
    By contrast, \sinvig-V0 is more redundant and sometimes asks repeated questions.
    }
    \label{fig:hri_example}
    \vspace{-15pt}
\end{figure*}

We have conducted extensive ablation studies to figure out what \sinvig learned from the self-generated data.
Concretely, the self-generated data mainly comprises two parts: 1) the extensive unlabeled image data; 2) LLM-polished visual disambiguation dialogues.
Therefore, we trained two extra models to figure out the contributions of each part, starting from \sinvig-R0.
These two models are further trained using images together with self-generated dialogues but without LLM polishing, namely \sinvig-R1-No-LLM and \sinvig-R2-No-LLM.
Results are demonstrated in \figref{fig:invig_exp}(b).
We can conclude that:
1) Training with data from LLM polishment consistently improves model performance.
2) Without the polishment of LLM, self-evolving may harm the performance, due to the low-quality data generated by the model itself, which harms the ability of text generation.
To intuitively demonstrate the data quality from \sinvig-R0 and \sinvig-R1, we have shown some examples in \figref{fig:data_r0_r1} in the \apref{ap:data_r0_r1}.

Besides, we also conducted cross-validation experiments to further verify the improvements of individual tasks.
Results are shown in \tabref{table:cross_val}.
We can see that after self-evolving, \sinvig iteratively improves the performance of each sub-module.
By comparing the performance of \sinvig-R0 combined with different Oracle models, it verifies the improvements of MT-VQA through self-evolving.
Similarly, by comparing the performance of \sinvig-R2 combined with different Guesser \& Questioner models, it verifies the improvements on MT-VQG and MT-VG.


\subsection{Does SInViG improve user experience by self-evolving?}
\label{sec:hri_exp}

To further validate the performance improvements from the self-evolving iteration in realistic human-robot interaction, we have conducted experiments on TiO-HRI bench \cite{xu2024towards} using \sinvig series.
TiO-HRI bench includes 150 images from InViG \cite{invig}, Object365 \cite{shao2019objects365}, and OpenImages \cite{openimages}.
It covers three subsets: \textit{Scene Understanding}, \textit{Human Understanding}, and \textit{Language Understanding}, each including 50 selected and manually designed challenging examples.
To ensure fairness of comparison and avoid bias, we adopt single-blind validation, meaning that the involved volunteers are not able to see the model name during the interaction, and all three models will be randomly re-ordered for each example.
We have recruited 6 volunteers in this phase, each interacting with three models based on a 25 subset.

Results are shown in \figref{fig:hri_exp}.
We can observe that in the aspect of success rate, \sinvig-R2 achieves the highest performance except for language understanding.
It is also preferred compared to the other two models during human evaluation.
From the survey after interaction experiments, we summarize the reasons that the volunteers prefer \sinvig-R2 as follows:
1) in most cases, it asks fewer questions and is more efficient;
2) it can ask more informative questions including richer semantic expressions.
We have also shown some examples of realistic human-robot interaction in \figref{fig:hri_example} to illustrate their differences.

\subsection{How is SInViG deployed on real-robot platforms?}
\label{sec:real_robot_exp}

\begin{figure*}[t!]
    \centering
    \includegraphics[width=\textwidth]{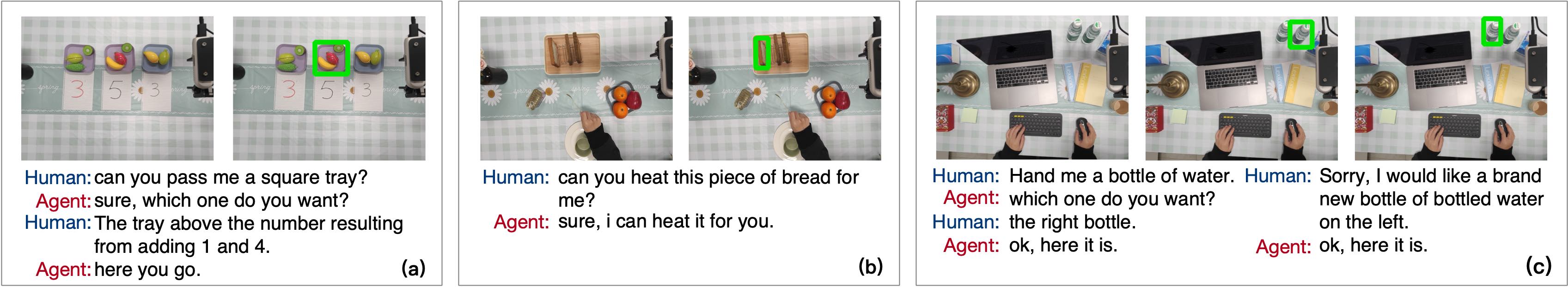}
    \caption{More examples of \sinvig in realistic applications. (a) Understanding simple calculations. (b) Multi-modal interaction. (c) Online correction.}
    \label{fig:extra_examples}
    \vspace{-20pt}
\end{figure*}

\begin{figure}
    \centering
    \includegraphics[width=0.9\columnwidth]{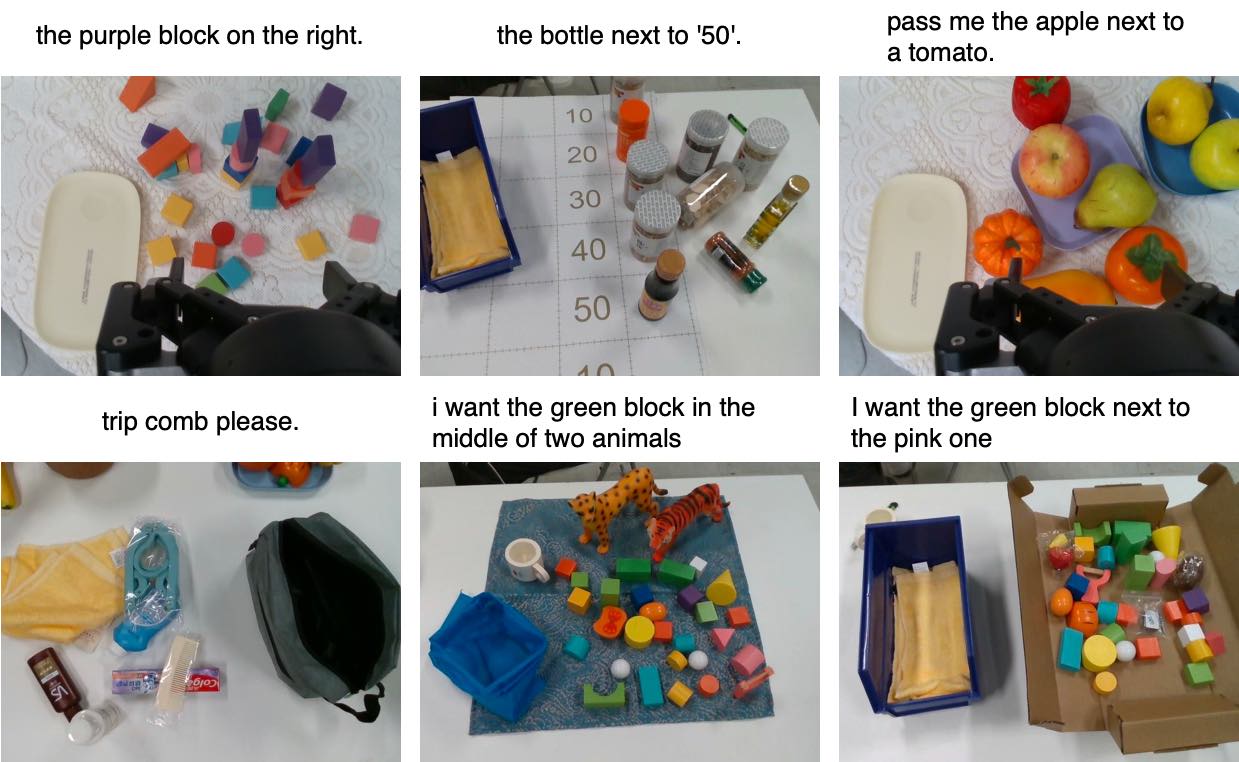}
    \caption{Scenarios for interactive manipulation tasks. Our real-robot scenes are designed to test the robustness against diversified visual and linguistic inputs.}
    \label{fig:real_robot_scenes}
    \vspace{-15pt}
\end{figure}

We have conducted real-robot experiments on 23 interactive manipulation scenarios.
Similar to human-robot interaction experiments, we recruit 7 volunteers for interaction.
Each volunteer does not know the exact model name that is speaking during the interaction.
We have demonstrated 6 out of the 23 scenes in \figref{fig:real_robot_scenes} along with the initial instruction.
The complete testing scenarios are listed in \apref{ap:real_robot_scenes_full}.
As a result, we have totally conducted 161 real-robot experiments for interactive manipulation tasks.

\begin{figure}[t!]
    \centering
    \includegraphics[width=\columnwidth]{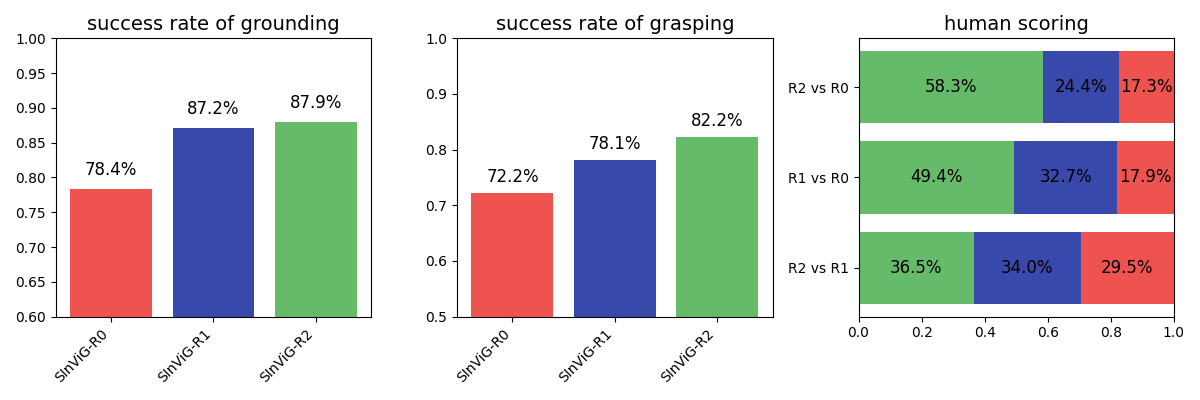}
    \caption{Results of real-robot experiments. Left: the success rate of interactive visual grounding; Middle: the success rate of interactive grasping; Right: the scores from human evaluation after interaction.}
    \label{fig:real_robot_exp}
    \vspace{-15pt}
\end{figure}

Results are shown in \figref{fig:real_robot_exp}.
Observations in the real-robot interactive tasks highly accord with human-robot interaction in \secref{sec:hri_exp}.
We can see that the success rate of \sinvig-R2 is higher than \sinvig-R0 with a clear margin.
Moreover, we can see from the human evaluation that self-evolving has substantially improved the scores of user experience.
Results here verified that our model can not only achieve state-of-the-art performance on interactive grounding benchmarks but also suffer no performance drop when deployed on the real-robot platform, with robustness against the complexity of real-robot scenarios.
We also combined \sinvig with state-of-the-art grasping algorithms \cite{sundermeyer2021contact} to test the compatibility of the output bounding boxes with physical manipulation tasks.
The results show that the output of \sinvig can be directly used for downstream grasping tasks without any further engineering.

Due to its robustness, simplicity, and flexibility, \sinvig can be deployed in a variety of practical scenarios.
We have listed three examples here.
\paragraph{Common Sence Understanding}
As shown in \figref{fig:extra_examples}(a), the evolved \sinvig can understand common sense during the interaction (e.g., character recognition and simple calculation), benefiting from the training using LLM-polished data.
It is evidence that our self-evolving iteration gradually distills the knowledge from the source LLM used for data polishing.

\paragraph{Multi-Modal Interaction}
\sinvig can interact with humans with multi-modal inputs, and understand instructions in the form of both languages and gestures.
It has the potential to be deployed for users with specific requirements.

\paragraph{Online Correction}
Due to its simplicity and extensive training in interaction data, \sinvig can accept additional descriptions even after the final decision has been made, as shown in \figref{fig:extra_examples}(d).
The additional descriptions can be directly concatenated after the dialogue history to correct the result if it is wrong.
This feedback can be useful when being deployed on a real robot as a way to prevent potential damage.

\section{Conclusions}
\label{sec:conclusion}


In this paper, we propose \sinvig, a self-evolving interactive visual grounding system, that can continuously and automatically learn from unlabeled images and large language models, without any intervention from humans.
We have conducted extensive experiments to verify the effectiveness and robustness of \sinvig, including the evaluation of interactive visual grounding benchmarks, human-robot interaction, and interactive robot manipulation tasks.
Our experimental results have demonstrated the advances of \sinvig over previous state-of-the-art models \cite{xu2024towards} with a clear margin.
Moreover, our results also verified that with the help of fully automated, self-evolving iterations, \sinvig improves user experience significantly by acquiring more and more preferences from human users during human-robot interaction experiments.
Our real-robot interactive manipulation tasks also present the practicality of \sinvig for diversified user demands.
It achieves robust performance against multi-modal interaction modes, open-ended instructions, and highly complex visual observations.

One known limitation of \sinvig is the occasional hallucination \cite{ji2023survey}.
We have observed that \sinvig occasionally gives ambiguous descriptions that do not match the contents of given images, especially when the visual instances in the images are in low resolution or severely partially observable.
We hypothesize that such hallucinations are inherited from large language models since we have observed that polished annotations from large language models sometimes include unmentioned contents in the original dialogues.
To alleviate such hallucinations, one possible way is to introduce human feedback for training a classifier or reward model \cite{ouyang2022training}.
Besides, on-body deployment of \sinvig, especially on mobile robotic platforms, poses challenges of changing perspectives during interaction.
Currently, \sinvig is training with images of static views and hence achieves limited performance in such scenarios with changing views.
Future work includes a straightforward extension with video inputs.




\bibliographystyle{plainnat}
\bibliography{references}

\newpage

\begin{table}
\caption{Training Hyper-parameters}
\label{table:hparam}
\begin{center}
\begin{tabularx}{1\columnwidth}{l@{\hspace*{65pt}}r@{}}
\toprule
 Param. Name & Value \\
\midrule
Batch Size & 14$\times$8 \\
Learning Rate & 3e-5 \\
Optimizer & Adam \\
Encoder Sequence Length & 1024 (Image) + 512 (Text) \\
Encoder Layers & 24 \\
Decoder Sequence Length & 256 \\
Decoder Layers & 12 \\
Hidden Size & 1280 \\
Intermediate Size & 5120 \\
Number of Attention Head & 16 \\
\bottomrule
\end{tabularx}
\end{center}
\end{table}

\begin{figure}[H]
    \centering
    \includegraphics[width=\columnwidth]{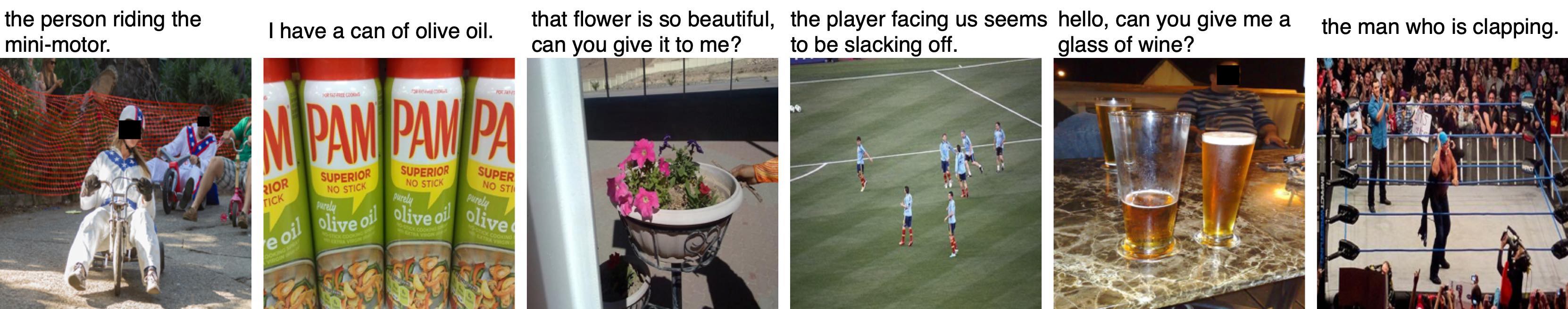}
    \caption{Examples removed from the HRI experiments compared to GPT-4V \cite{achiam2023gpt} and Gemini-Pro \cite{team2023gemini} due to their rejection to interact based on these images.}
    \label{fig:invalid_hri_exampels}
\end{figure}

\begin{table}[H]
\caption{Comparison to Multi-Modal LLMs}
\label{table:llm_exp}
\begin{center}
\begin{tabularx}{1\columnwidth}{l@{\hspace*{5pt}}|l@{\hspace*{5pt}}|l@{\hspace*{20pt}}|c@{\hspace*{10pt}}|c}
\toprule
  \specialrule{0em}{1pt}{1pt}
Questioner & Guesser & Oracle & SR & Avg. Latency\\
  \specialrule{0em}{1pt}{1pt}
\hline
  \specialrule{0em}{1pt}{1pt}
\multicolumn{2}{c|}{GPT-4V \cite{achiam2023gpt}} & Human & $<$5\% & 9.2s \\
  \specialrule{0em}{1pt}{1pt}
\hline
  \specialrule{0em}{1pt}{1pt}
\multicolumn{2}{c|}{Gemini-Pro \cite{team2023gemini}} & Human & $<$5\%& 13.7s \\
  \specialrule{0em}{1pt}{1pt}
\hline
  \specialrule{0em}{1pt}{1pt}
GPT-4V \cite{achiam2023gpt} & \sinvig-R2 & Human & 70.9\% & 9.2s \\
  \specialrule{0em}{1pt}{1pt}
\hline
  \specialrule{0em}{1pt}{1pt}
Gemini-Pro \cite{team2023gemini} & \sinvig-R2 & Human & 70.9\% & 13.7s \\
  \specialrule{0em}{1pt}{1pt}
\hline
  \specialrule{0em}{1pt}{1pt}
\multicolumn{2}{c|}{\sinvig-R2} & Human & 75.4\% & 0.4s \\

\bottomrule
\end{tabularx}
\end{center}
\end{table}

\begin{figure}[H]
    \centering
    \includegraphics[width=\columnwidth]{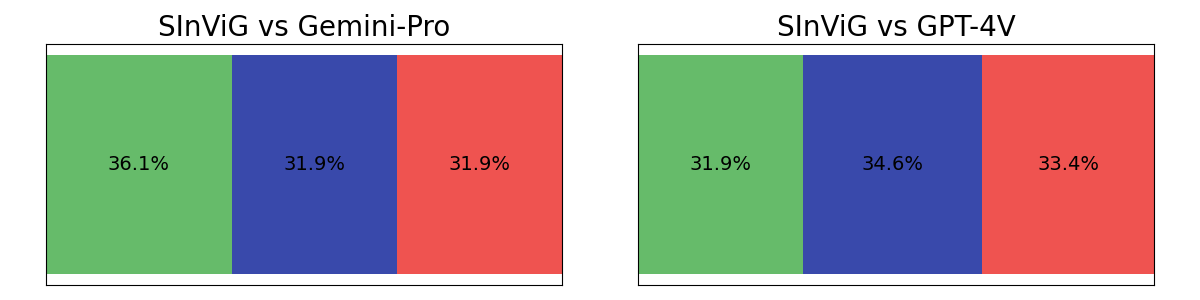}
    \caption{Human evaluation scores for comparison between \sinvig and commercial multi-modal large language models: Gemini-Pro (Left) \cite{team2023gemini} and GPT-4V (Right) \cite{zhu2023minigpt4}.}
    \label{fig:human_scores_llm}
\end{figure}

\section*{Appendix}
\subsection{Training Hyperparams}
\label{ap:hyperparam}

We listed the detailed training hyper-parameters in \tabref{table:hparam} for reference.

\subsection{Compaison to GPT-4V and Gemini-Pro}
\label{ap:mllms_comp}

To comprehensively understand the performance of \sinvig, we have conducted experiments to compare \sinvig with state-of-the-art multi-modal LLMs including GPT-4V \cite{achiam2023gpt} and Gemini-Pro \cite{team2023gemini}, which are two well-known LLMs for commercial usage.
The evaluation is conducted on the TiO-HRI bench \cite{xu2024towards} as in \secref{sec:hri_exp}, with 13 recruited volunteers as the users.
Totally, we conducted 421 HRI experiments, each including the comparison of all three models.
As in \secref{sec:hri_exp}, all volunteers are not able to see the model name during the interaction.
Noticeably, we find that some examples are not accepted by GPT-4V or Gemini-Pro due to unknown reasons, we listed them in \figref{fig:invalid_hri_exampels}.
Hence, in this section, all performances are based on the rest images.

Results are shown in \tabref{table:llm_exp}.
Since both GPT-4V and Gemini-Pro are not designed for vision tasks like visual grounding, they can hardly output the location of objects directly.
We tried our best to make it work.
Nevertheless, the interactive visual grounding performance is less than 5\% overall without the help of visual grounding models.
Therefore, we introduce a much stronger baseline, in which we use our best model, SInViG-R2, as the Guesser and GPT-4V or Gemini-Pro as the Questioner.
Under this setting, the comparison is totally on the multi-modal interaction.
We can conclude that with the help of \sinvig-R2, both GPT-4V and Gemini-Pro can achieve commendable performance for the interactive visual grounding tasks.
Nevertheless, their latency is somewhat unacceptable when deployed on real-time robot systems.
We also demonstrate the human preference in \figref{fig:human_scores_llm}.
We can see that without knowing the specific model ID, humans prefer the GPT-4V the most, benefiting from its powerful multi-modal interaction ability.
Nevertheless, our model, with much fewer training data and a smaller model size, is more preferred compared to Gemini-Pro.

\begin{figure}[t]
    \centering
    \includegraphics[width=0.85\columnwidth]{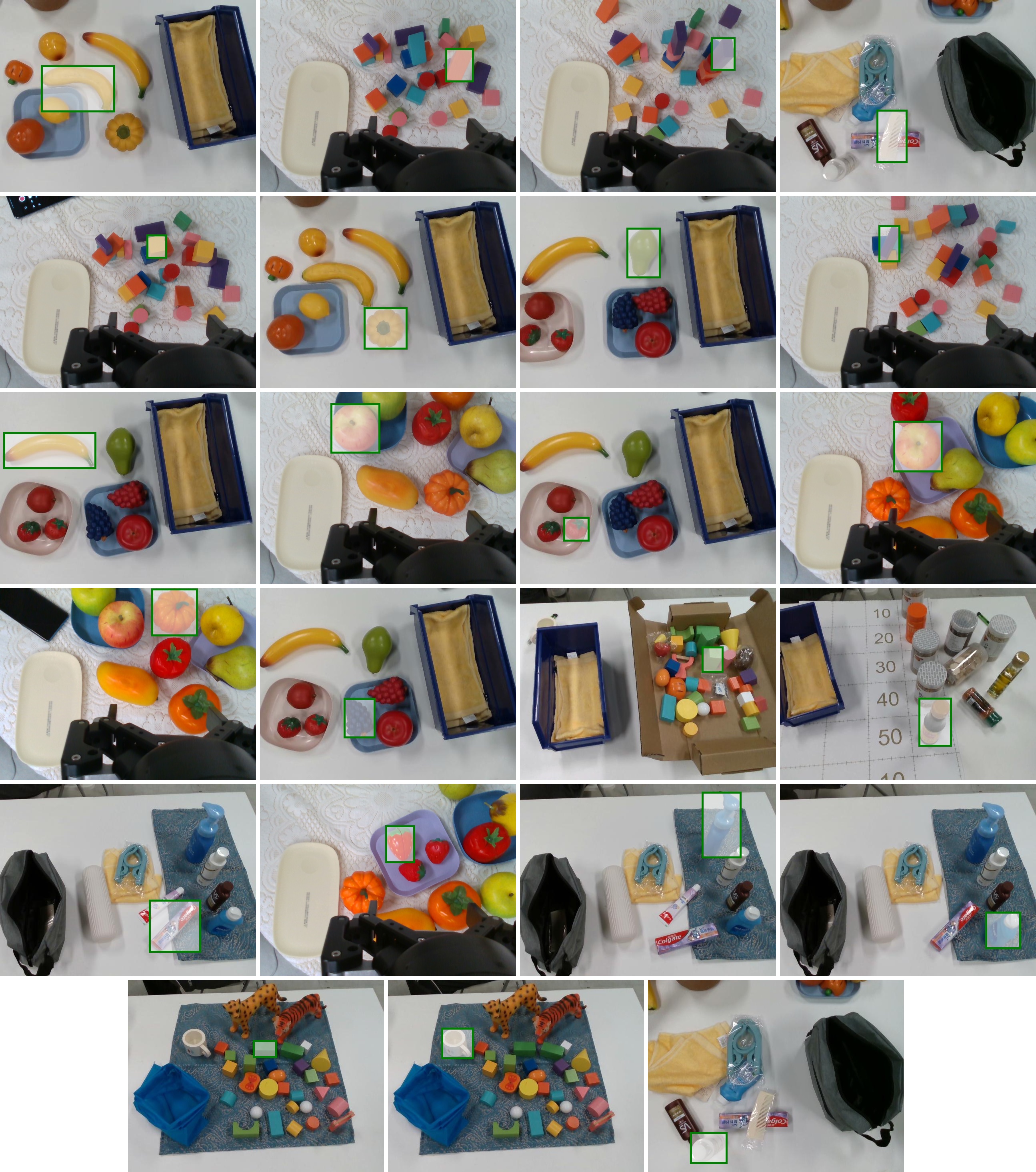}
    \caption{Experimental scenes of real-robot interactive manipulation tasks. The green bounding box shown in each image indicates the target object.}
    \label{fig:robot_exp_scenes_full}
\end{figure}

\begin{table*}[t!]
\caption{Detailed Performance on InViG Benchmark}
\label{table:performance_details}
\begin{center}
\begin{tabularx}{1\textwidth}{l@{\hspace*{3.5pt}}c@{\hspace*{3.5pt}}c@{\hspace*{3.5pt}}c@{\hspace*{3.5pt}}c@{\hspace*{3.5pt}}c@{\hspace*{3.5pt}}c@{\hspace*{3.5pt}}c@{\hspace*{3.5pt}}c@{\hspace*{3.5pt}}c@{\hspace*{3.5pt}}c@{\hspace*{3.5pt}}c@{\hspace*{3.5pt}}c@{\hspace*{3.5pt}}c@{\hspace*{3.5pt}}c@{\hspace*{3.5pt}}c@{\hspace*{3.5pt}}c@{\hspace*{3.5pt}}c@{\hspace*{3.5pt}}c@{\hspace*{3.5pt}}c@{\hspace*{3.5pt}}c@{}}
\toprule
 & \multicolumn{9}{c}{MT-VQG} & \multicolumn{9}{c}{MT-VQA} & \multirow{2}{*}{MT-VG} & \multirow{2}{*}{IVG} \\
\specialrule{0em}{1pt}{1pt}
\cmidrule(r){2-10} \cmidrule(r){11-19}
& B1 & B4 & R & M & C & S & R@1 & R@5 & Rank &  B1 & B4 & R & M & C & S & R@1 & R@5 & Rank \\
\midrule
\sinvig-HO-R0 & \bf 0.382 & 0.218 & 0.385 & 0.200 & 1.812 & 0.243 & 46.7\% & 72.6\% & 4.8 &  0.257 &0.116& 0.325 & 0.166 & 1.097 &\bf 0.223& 50.2\% & 71.6\% & 5.4 & 71.7\% & 37.1\%\\
\specialrule{0em}{1pt}{1pt}
\sinvig-HO-R1 & 0.372 & 0.216 & 0.375 & 0.200 & 1.822 &0.237& 49.1\% & 75.6\% & 4.4 &0.277&0.127& 0.332 & 0.174 & 1.154 &0.218& 54.6\% & 76.3\% &4.6 & 73.9\% & 47.1\%\\
\specialrule{0em}{1pt}{1pt}
\sinvig-HO-R2 & 0.374 & \bf 0.220 & \bf 0.388 & \bf 0.210 & \bf 1.939 &\bf 0.246& \bf 51.4\% & \bf 78.2\% & \bf 4.0 &\bf 0.284&\bf 0.129& \bf 0.337 & \bf 0.176 & \bf 1.163 &0.221& \bf 56.8\% & \bf 78.3\% & \bf 4.3 & \bf 75.8\% &\bf 64.3\%\\
\midrule
\specialrule{0em}{1pt}{1pt}
\sinvig-R0 & 0.360 & 0.209 & 0.372&0.199&\bf  1.834 &\bf 0.240&47.4\% & 74.7\% & 4.7 & \bf 0.297 & \bf 0.129 &0.322 & \bf 0.176 & \bf 1.132 &\bf 0.236& 51.8\% & 72.8\% & 5.6 &76.8\%& 73.6\%\\
\specialrule{0em}{1pt}{1pt}
\sinvig-R1 & 0.361 & 0.202 &0.372&0.199& 1.757&\bf 0.240& 48.5\% & 75.4\% & 4.5 & 0.276 & 0.114 &0.312 & 0.165 &1.050 &0.205& 54.1\% & 75.2\% & 4.9 &77.5\%& 73.5\% \\
\specialrule{0em}{1pt}{1pt}
\sinvig-R2 &\bf  0.370 & \bf 0.215 &\bf 0.381&\bf 0.203&1.821&0.237& \bf 49.7\% & \bf 76.5\% &\bf  4.3 & 0.267 & 0.117 & \bf 0.329 & 0.170 & 1.127 &0.216& \bf 55.1\% & \bf 76.0\% & \bf 4.7 &\bf 77.4\%& \bf 80.9\%\\
\bottomrule
\end{tabularx}
\end{center}
\end{table*}

\subsection{Detailed Settings for Human-Robot Interaction}

\begin{figure}
    \centering
    \includegraphics[width=\columnwidth]{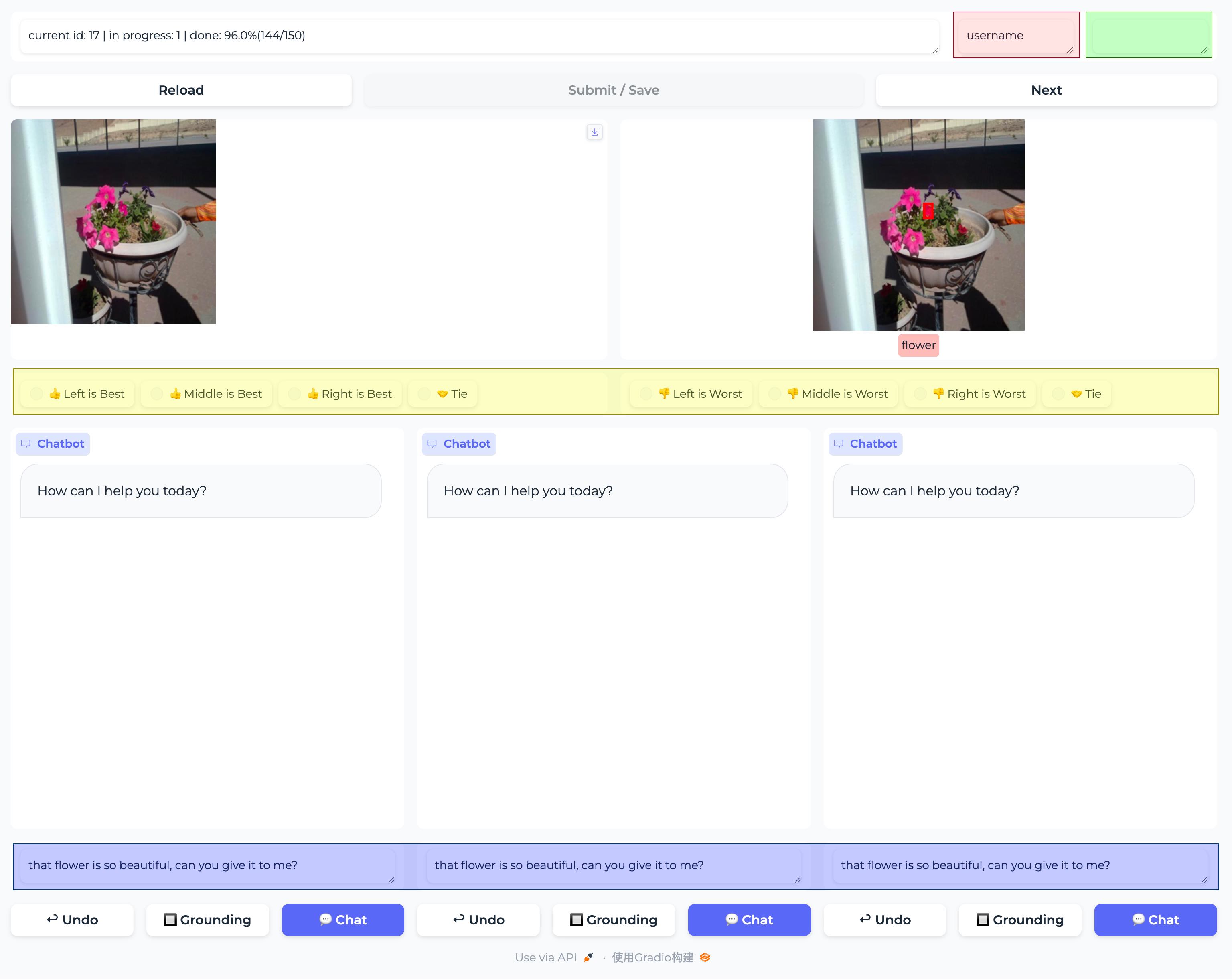}
    \caption{Interface used during all human-robot interaction experiments.}
    \label{fig:hri_exp_interface}
\end{figure}

In our human-robot interaction experiments, we design the interface for blind interaction as shown in \figref{fig:hri_exp_interface}, i.e., the user does not know the specific model IDs while interacting with them side by side.
This interface aims to lower the bias from users for human-robot interaction and make the results as fair as possible.

Concretely, our interaction web demo is based on Gradio \cite{abid2019gradio}, which is for testing models of Machine Learning research in the wild.
In the TiO-HRI bench \cite{xu2024towards}, all 150 examples are paired with a predefined initial instruction and a target object for standardization.
The blank on the top-right of the page marked in the red box is used to record the user names.
The blank in green is used to jump to a specific example, which will not be used unless errors are encountered.
Images, instructions, and target locations are listed in the middle.
Users are required to interact with three models, which will be loaded in a random order for each experiment, using the chat box marked in blue.
User scores will be collected after interaction using the buttons in the yellow box.
The logic of human evaluation is:
\begin{itemize}
    \item two ties means that all three models are equally good or bad;
    \item a tie (e.g., A and B) and a worst (e.g. C) mean that A is equal to B, and both are better than C;
    \item a tie (e.g., A and B) and a best (e.g. C) mean that A is equal to B, and both are worse than C;
    \item a best (e.g. A) and a worst (e.g. C) mean that A is better than B, which is better than C.
\end{itemize}

\begin{figure*}[t!]
    \centering
    \includegraphics[width=\textwidth]{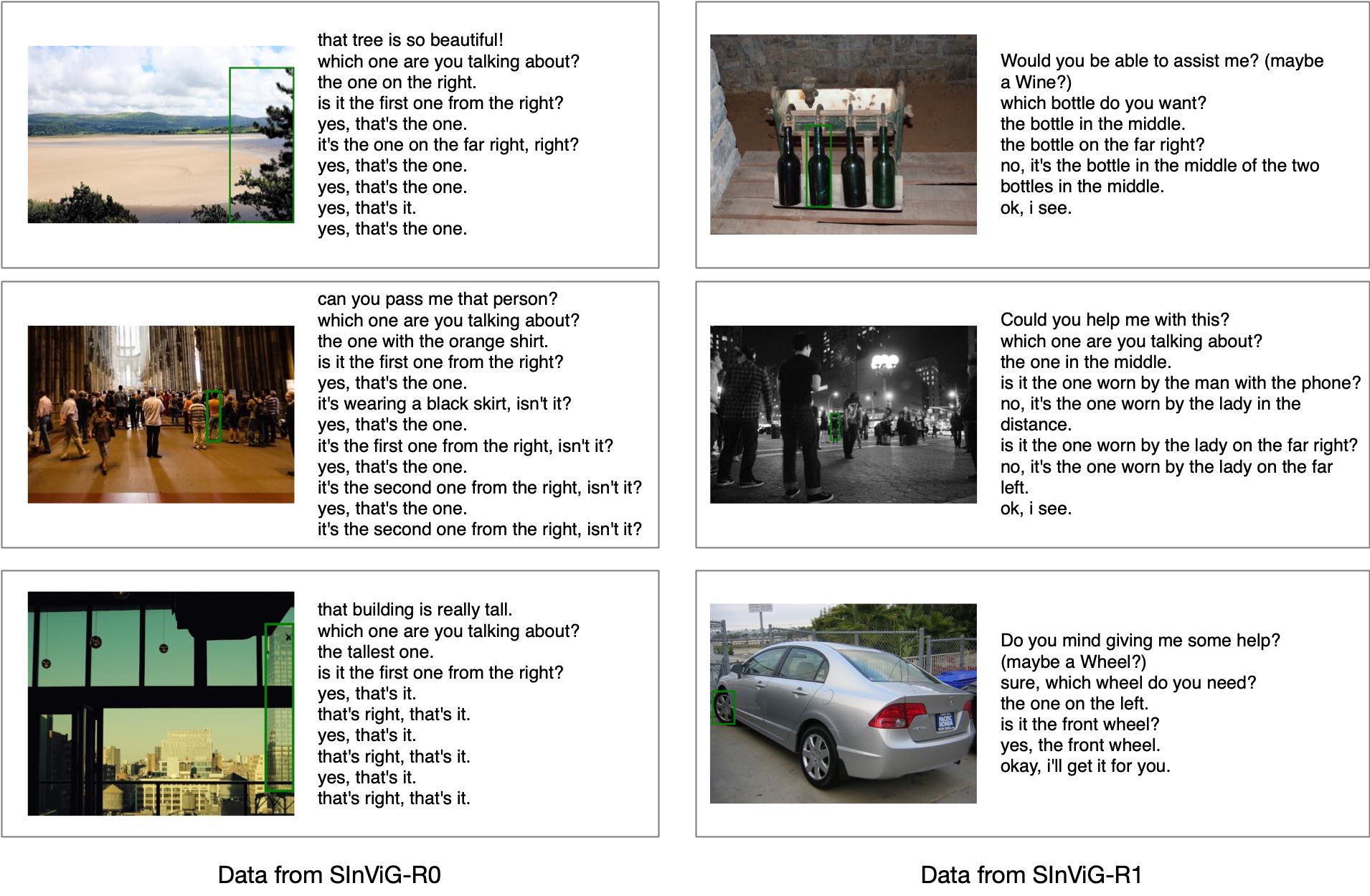}
    \caption{Generated data using different models: \sinvig-R0 (left) versus \sinvig-R1 (right).}
    \label{fig:data_r0_r1}
\end{figure*}

\subsection{Real-Robot Experimental Scenes}
\label{ap:real_robot_scenes_full}

We have listed all experimental scenes used during our real-robot interactive manipulation tasks in \figref{fig:robot_exp_scenes_full}.
In each scene, we specify a target object, which is marked as the green bounding boxes in \figref{fig:robot_exp_scenes_full}.
In our experiments, each volunteer is required to interact with the robot to pick the target object with unrestricted descriptions and interaction.
We record the success rate for each volunteer.
After the interaction experiments, we also asked them to score the models only with the model outputs without access to the model IDs.

\subsection{Data Generated using Different Models}
\label{ap:data_r0_r1}

We have shown some examples generated using \sinvig-R0 and \sinvig-R1 in \figref{fig:data_r0_r1}.
We can see that data generated using \sinvig-R0 is somewhat dirty, often with repeated sentences.
Besides, the distribution of the output words is narrow: it likes asking questions using spatial relationships instead of self-referential attributes.
By contrast, the data generated using \sinvig-R1 seems much better, with richer expressions and fewer repeated questions.

\subsection{Detailed Performance Comparison}
\label{ap:txt_sim}

We have shown the detailed performance of \sinvig here in \tabref{table:performance_details}.
We can see that for both series of \sinvig-HO and \sinvig, most metrics are improved after self-evolution.
Besides, we can find that \sinvig-HO evolves faster than \sinvig.
\sinvig-HO-R2 even achieves better performance on InViG benchmark when compared to \sinvig-R2, which is trained additionally on extensive publicly available vision-language datasets.
It reveals that self-evolving training can boost performance especially when there is no much available training data.

\end{document}